\documentclass[atmp]{ipart_v1}
\Vol{NN}
\Issue{1}
\Year{2025}
\firstpage{701}

\usepackage{t1enc}
\usepackage[latin1]{inputenc}
\usepackage[english]{babel}
\usepackage{multirow}
\usepackage{placeins}
\usepackage{subcaption}
\usepackage{graphicx} 

\usepackage{amsthm}
\usepackage{yfonts}

\usepackage{bbm}
\usepackage{bm}
\usepackage{mathrsfs}
\usepackage{circuitikz}
\ifatmp
\usepackage{authblk}
\fi
\usepackage{listings}
\usepackage{caption}
\usepackage{courier}
\usepackage{hyperref}

\definecolor{mylinkcolor}{rgb}{0.5,0.0,0.0}
\definecolor{myurlcolor}{rgb}{0.0,0.0,0.5}
\definecolor{mycitecolor}{rgb}{0.0,0.5,0.0}
\hypersetup{colorlinks=true,urlcolor=myurlcolor,citecolor=mycitecolor,linkcolor=mylinkcolor,linktoc=page,breaklinks=true}

\newcommand{\DHL}{\mathrm{DHL}}

\newcommand{\be}[0]{\begin{equation}}
\newcommand{\ee}[0]{\end{equation}}

\numberwithin{equation}{section}

\theoremstyle{plain}

\title[Generative Modeling for Mathematical Discovery]{Generative Modeling for Mathematical Discovery}

\author[a]{Jordan S. Ellenberg}
\author[b]{Cristofero S. Fraser-Taliente}
\author[c,d]{Thomas R. Harvey}
\author[a]{Karan Srivastava}
\author[c]{Andrew V. Sutherland}
\affil[a]{University of Wisconsin-Madison}
\affil[b]{University of Oxford}
\affil[c]{Massachusetts Institute of Technology}
\affil[d]{The NSF Institute for Artificial Intelligence and Fundamental Interactions}

\begin{document}

\begin{abstract}
We present a new implementation of the LLM-driven genetic algorithm {\it funsearch}, whose aim is to generate examples of interest to mathematicians and which has already had some success in problems in extremal combinatorics.  Our implementation is designed to be useful in practice for working mathematicians; it does not require expertise in machine learning or access to high-performance computing resources.  Applying {\it funsearch} to a new problem involves modifying a small segment of Python code and selecting a large language model (LLM) from one of many third-party providers. 
We benchmarked our implementation on three different problems, obtaining metrics that may inform applications of {\it funsearch} to new problems.
Our results demonstrate that {\it funsearch} successfully learns in a variety of combinatorial and number-theoretic settings, and in some contexts learns principles that generalize beyond the problem originally trained on.
\end{abstract}

\maketitle

\section{Introduction}

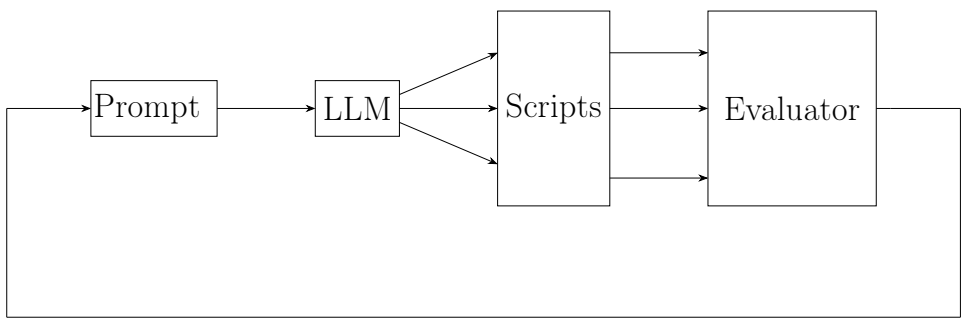
\begin{figure}[!ht]
\centering
\resizebox{1\textwidth}{!}{%
\begin{circuitikz}
\tikzstyle{every node}=[font=\LARGE]
\node [font=\LARGE] at (1.25,8.25) {Prompt};
\node [font=\LARGE] at (5,8.25) {LLM};
\node [font=\LARGE] at (8.5,8.25) {Scripts};
\node [font=\LARGE] at (12.75,8.25) {Evaluator};
\draw [->, >=Stealth] (5.75,8.25) -- (7.5,8.25);
\draw [->, >=Stealth] (9.5,8.25) -- (11.25,8.25);
\draw [->, >=Stealth] (5.75,8.5) -- (7.5,9.25);
\draw [->, >=Stealth] (5.75,8) -- (7.5,7.25);
\draw [->, >=Stealth] (9.5,9.25) -- (11.25,9.25);
\draw [->, >=Stealth] (9.5,7) -- (11.25,7);
\draw (14.5,8.25) to[short] (15.75,8.25);
\draw (15.75,8.25) to[short] (15.75,4.5);
\draw  (5.75,7.75) rectangle (4.25,8.75);
\draw  (2.5,7.75) rectangle (0.25,8.75);
\draw [->, >=Stealth] (2.5,8.25) -- (4.25,8.25);
\draw  (7.5,10) rectangle (9.5,6.5);
\draw  (11.25,10) rectangle (14.25,6.5);
\draw [short] (14.5,8.25) -- (14.25,8.25);
\draw [->, >=Stealth] (-1.25,8.25) -- (0.25,8.25);
\draw [short] (15.75,4.5) -- (-1.25,4.5);
\draw [short] (-1.25,8.25) -- (-1.25,4.5);
\end{circuitikz}
}%
\caption{The basic structure of \textit{funsearch}.}
\label{fig:funsearch}
\end{figure}

\textit{Funsearch} is a computational method for mathematical discovery that exploits the apparent coding expertise of large language models~(LLMs)~\cite{romera2024mathematical}. The structure of \textit{funsearch} is that of a genetic algorithm, where the population is a large collection of candidate Python scripts. These scripts are scored by an external evaluator, and future generations of scripts are generated by inputting some of the previous generation's scripts into the prompts of an LLM. The scripts selected for constructing the prompts are chosen based on their scores, while ensuring diversity within the overall population to avoid convergence to local minima. This basic structure is shown in Figure~\ref{fig:funsearch}, and we proceed to describe each stage in further detail, but more exact specifications can be found in the appendices of the original paper~\cite{romera2024mathematical}.

\textit{funsearch} is most effectively applied to inverse problems, where it is relatively easy (typically solvable in polynomial time) to verify or score a proposed solution, but significantly more challenging (often requiring super-polynomial time) to discover such a solution in the first place. This approach is particularly suited to combinatorial optimization problems, such as the cap set problem introduced in Section~\ref{sec:capset}. In practice, the authors of \cite{romera2024mathematical} discovered that the most effective results are obtained by formulating the problem using a priority function. In the simplest instance, the priority function dictates the order in which elements should be assembled from a larger set to create a special subset. A deterministic function ensures that the special subset is always valid by preventing invalid elements from being added. The priority function should therefore be thought of as a heuristic guiding rigorous subset selection. It is the opinion of the authors that \emph{funsearch} is likely to be most useful in such `hard-soft' settings, where the program designed by the language model is allowed to be `imprecise' in some appropriate sense, because it is monitored by a rigorous evaluation function. As an example implementation, Appendix \ref{app:ExampleSpec} provides a straightforward \textit{funsearch} specification file for determining whether a number is prime\footnote{This example turns out to be too simple for useful benchmarking, where the correct program is frequently found within the first generation; it is worth noting that \textit{funsearch} has much more trouble when asked to determine whether a number is five less than a prime, a question which is formally of the same level of complexity.}.

The specification includes three functions: {\it evaluate}, {\it solve}, and {\it priority}.\footnote{For clarity, we will always call the evolving function the `priority' function.} The {\it evaluate} function assesses and scores a given proposed solution. A proposed solution consists of both the {\it solve} and {\it priority} functions. \textit{funsearch} evolves a population of {\it priority} functions, keeping the {\it solve} function constant, in an effort to maximize the score provided by the {\it evaluate} function.  The distinction between {\it priority} and {\it solve} can best be understood by an example.  Suppose, for instance, we were trying to construct a large graph on $N$ vertices with no $4$-cycles.  The {\it priority} function might be a function that assigns a real number to each pair of vertices, which we think of as a potential edge.  Then we might define the {\em solve} function as follows:
\begin{itemize}
    \item start with the empty graph;
    \item check whether the highest-priority edge among those not only in the graph would create a $4$-cycle if added;
    \item if so, add it; if not, move on to the next highest-priority edge;
    \item if there are no more edges you can add without creating a $4$-cycle, terminate and return the current $4$-cycle-free graph.
\end{itemize}

In other words, this version of {\it solve} is a greedy algorithm and {\it priority} tells us in what order we should present the greedy algorithm with the edges.  (In this case, {\it evaluate} would just return the number of edges in the graph returned by {\it solve}.)

There are many other ways we could set up {\it solve}.  For instance, we could have {\it solve} terminate the first time the highest-priority edge would create a $4$-cycle, instead of skipping on to the next highest-priority edge.  Or {\it solve} could start with the complete graph and {\it remove} edges in order of priority, skipping any edges that aren't contained in a $4$-cycle, terminating when there are no more $4$-cycles left to remove.  We could also define {\em priority} to be a function which takes as input a graph and a pair of vertices, rather than just a pair of vertices; then {\em solve} could take into account the current graph $\Gamma$, calling {\it priority}$(\Gamma, \mbox{edge})$ in order to determine the priority of an edge given~$\Gamma$.

A {\it priority} function is not intrinsically good or bad for a given problem; it is only good or bad with reference to a specific choice of {\it solve}, which is always held constant over the course of evolution.

The utility of the separation of {\it priority} and {\it solve} is that {\it solve} is under the full control of the human operator, so if there are constraints we wish to impose on the solutions (for instance, that they obey a certain symmetry) we can accomplish that by putting a restriction in {\it solve}.  The {\it priority} function, by contrast, is the one evolved by the LLM; we have very little control over its features other than information provided in comments or the docstring of the function that will be included in the prompt to the LLM.

The population of {\it priority} functions consists of valid Python scripts, which are evolved according to an island model from genetic algorithms~\cite{cantu1998survey,goldberg1989genetic}. The population is split into sub-populations, called islands, which evolve independently. For selection, an island is chosen first, followed by the random selection of a program within that island weighted by its score. We periodically discard the half of the islands where the highest score is least, and reinitialise these islands with scripts from the remaining islands.

By the above sample procedure, we select $N$ scripts from a single island to create the prompt for a pre-trained LLM\footnote{We, and the authors of~\cite{romera2024mathematical}, set $n=2$.}. These $n$ scripts are sorted by their score, and the $k$th entry is renamed to {\it``priority\_v$\langle k-1 \rangle$''}. The prompt is then of the same format as the specification (see Appendix~\cite{romera2024mathematical}), but with all $N$ priority functions listed along with the header for {\it priority\_vN}. This prompt is then fed into the LLM, which generates a suggestion for a new potential priority function. This stage is akin to the breeding and mutation phases of a traditional genetic algorithm.

\section{Implementation}

Our implementation of {\it funsearch} is available in the GitHub repository~\cite{funsearch}

\begin{center}
\url{https://github.com/kitft/funsearch}
\end{center}

One can easily apply \textit{funsearch} to a problem by writing a configuration file such as that in Appendix~\ref{app:ExampleSpec}. Other configuration files can be found in the \texttt{examples} folder of the GitHub repository \cite{funsearch}; this includes the three problems we used in our benchmarks, which are discussed in the next section. We introduce a number of new features not available in the original implementation of \textit{funsearch} \cite{romera2024mathematical}:
\begin{itemize}
    \item support for non-priority-function program search (i.e. searching for a function with arbitrary type signature);
    \item integration with Weights \& Biases \cite{wandb} for real-time monitoring and logging;
    \item support for multiple LLM platforms or OpenRouter, as well as mixing models within runs;
    \item parallel processing: produced functions are evaluated across multiple CPU cores and asynchronous API calls;
    \item improved safety features: code produced by the LLM is checked before runtime for potentially dangerous operations (e.g. calls to the shell), and sandboxing is supported at both the process and container level.
\end{itemize}

OpenRouter is a convenient service that provides API access to a variety of large language models from different providers. It acts as a routing layer; we can use a single API endpoint instead of having to separately integrate each provider's API. This simplifies the process of experimenting with different LLMs in the \textit{funsearch} implementation.

\section{Problems}

In this section we present the three problems we used to test our new implementation of {\it funsearch}: cap-set, narrow-admissible-tuple, and no-isosceles.

\subsection{Cap sets}\label{sec:capset}
The \textit{cap-set} problem~\cite{grochow2019new} was the original problem considered for \textit{funsearch} in~\cite{romera2024mathematical}, which found the largest known cap set of size 512 in 8 dimensions. As such, it makes sense to include this problem. However, we must acknowledge that any model trained after the release of the original \textit{funsearch} article may have incorporated the record-breaking program into its training data. This presents a potential issue that is likely to become a general concern in the future when benchmarking LLMs on mathematical tasks.

The baseline {\it priority--solve} setup for cap-set is as follows.
\begin{itemize}
    \item {\it priority} takes as input an integer $n$ and an element of $(\mathbf{Z}/3\mathbf{Z})^n$, and outputs a real number.
    \item {\it solve} takes $n$ as input, starts with an empty subset of $(\mathbf{Z}/3\mathbf{Z})^n$ and adds points in order of {\it priority}, skipping any points that would create a set of three vectors summing to zero.  Terminate when no more points can be added without violating this constraint.  The {\it evaluate} function returns the size of the resulting set.
\end{itemize}

The cap-set problem is one of combinatorial optimization: the task is to find the largest possible set of vectors in $\mathbb Z_3^n$ such that no three vectors sum to zero. 
Exact results are known up to $n=6$, but the exponential growth of the environment with $n$ quickly leads exhaustive searches intractable. At $n=8$, the environment consists of roughly $10^{760}$ combinations.

\subsection{Narrow admissible tuples}

An increasing sequence $t$ of $k$ integers that do not occupy every residue class modulo any prime $p$ is called an \textit{admissible $k$-tuple}. If $t=(t_1,\ldots,t_k)$ is an admissible $k$-tuple, then so are its translates $t+a=(t_1+a,\ldots,t_k+a)$ by integers $a$, so we assume without loss of generality that $t_1=0$.  Examples of admissible $k$-tuples are $(0,2)$, $(0,2,6)$, and $(0,4,6)$, while $(0,1)$ and $(0,2,4)$ are nonexamples.  The \textit{diameter} of an admissible $k$-tuple $(t_1,\ldots,t_k)$ is $t_k-t_1$.

For positive integers $j\le k$ the Dickson--Hardy--Littlewood conjecture $\DHL[k,j]$ states that every admissible $k$-tuple has infinitely many translates that contain at least $j$-primes;  $\DHL[2,2]$ implies the twin prime conjecture via the admissible $2$-tuple $(0,2)$, a conjecture that remains open.  Yitang Zhang \cite{zhang2014} proved that $\DHL[3500000,2]$ holds, which was eventually improved by the Polymath8 \cite{polymath8b} project on prime gaps to $\DHL[50,2]$.  James Maynard \cite{maynard2015} (and independently Terence Tao) proved that for every positive $j$ there exists a $k$ for which $\DHL[k,j]$ holds, and the Polymath8 project proved $\DHL[k,j]$ for $j=2,3,4,5,6$ and explicit values of $k$, including $k=35410$ for $j=3$.  Julia Stadlmann \cite{stadlmann2023} recently proved $\DHL[k=35265,3]$, which is currently the best known result for $j=3$.

Once $\DHL[k,j]$ is known, every admissible $k$-tuple of diameter $d$ implies the existence of infinitely many intervals of width $d$ that contain $j$ primes. This motivates the \textit{narrow-admissible-tuple} problem: given an integer $k$, find an admissible $k$-tuple whose diameter is as small as possible.

Minimal diameters are known for all $k\le 342$, and for $k=50$ this optimal diameter is $242$, which implies that there are infinitely many pairs of adjacent primes separated by a gap of at most 242.
For $k=35265$ the narrowest admissible $k$-tuple known has diameter $d=396504$, but this diameter is almost surely not minimal.  The discovery of any $35265$-tuple of smaller diameter would improve the least $d$ for which one can prove that there are infinitely many triples of primes separated by a gap of at most $d$.

For $k\le 5000$ examples of the narrowest known admissible $k$-tuples can be found at \cite{tuples}. This database was constructed as part of the Polymath8 project using a variety of methods \cite{natalg}, including some genetic algorithms.  While many CPU-years of computation were invested in the construction of this database, it remains likely that many of these tuples can be improved, especially for larger values of $k$.  For $k=35265$ and $d\approx 396504$ there are more than $10^{50000}$ $k$-tuples whose admissibility one might want to consider.

To search for narrow admissible tuples using \textit{funsearch}, rather than fixing~$k$, we fix a bound $n$ on the diameter and construct an admissible tuple by sieving the integers in $[0,n]$, removing one residue class for each prime $p\le n$.  The $k$ integers that remain will form an admissible $k$-tuple of diameter at most $n$.  The priority function takes integer inputs $p$ and $n$, with $p\le n$ prime, and returns an integer $r$ specifying the residue class $r+p\mathbb Z$ to remove. The starting {\it priority} function simply returns $1$ on all inputs.

We note that with this setup the starting {\it priority} function and all constant {\it priority} functions will yield only an admissible 2-tuple at best, so there is a strong selection pressure toward nonconstant priority functions.  We view this as a feature for the purpose of testing the evolutionary capability of {\it funsearch}, although it may make it more difficult to obtain record-breaking results.  An alternative approach along the lines of cap-set would have the {\it priority} function rank the integers in $[0,n]$ while the {\it solve} function builds an admissible tuple by adding integers in order by rank, skipping those that violate the admissibility condition.  With this alternative setup a constant priority function would already produce a fairly narrow admissible tuple that is close to optimal for small values of $n$.

\subsection{Subsets of the grid with no isosceles triangles}

We say that a subset $S$ of the regular $n \times n$ grid is {\em no-isosceles} if no three distinct points form an isosceles triangle.  (This constraint includes {\em collinear} triples, in which an isosceles triangle is just a three-term arithmetic progression.)  The problem of producing a large no-isosceles subset is an appealing one for machine learning methods.  As with the capset problem, the difference between the best known upper and lower bounds is very large.  The best known lower bound is on the order of $n/\sqrt{\log n }$~\cite[Appendix A]{patternboost}, while for upper bounds the best we can do is $n^{2-\epsilon}$, arising from the easy observation that the intersection of $S$ with any column of the grid is a subset of the interval with no three-term arithmetic progression.  Applying \textit{funsearch} to this problem allows for direct comparison with another transformer-based example generation protocol, PatternBoost~\cite[\S 4.1]{patternboost}.  When $n=64$, the largest set of points found by any of our models in our benchmark has size $96$; by contrast, PatternBoost found a no-isosceles set of size $110$ (starting from a large population of sets of size $108$ generated by non-LLM methods.)

We note that the no-isosceles problem, like the well-known no-three-in-line problem, asks for large subsets of the grid that do not contain a triple of points $(x_1,y_1),(x_2,y_2),(x_3,y_3)$ for which a certain quadratic polynomial $Q(x_1,y_1,x_2,y_2,x_3,y_3)$ vanishes.  Automated approaches to these two problems and other problems in this family will likely be valuable in understanding what happens for general quadratic forms $Q$.

By contrast with the other two problems discussed here, our main focus in no-isosceles will not be comparing different models; beyond the baseline benchmark, the work on no-isosceles here is centered on comparing different ways of presenting \textit{funsearch} with the same combinatorial problem, and about how \textit{funsearch} behaves when presented with variants of the problem.  A crucial thing mathematicians need to keep in mind when working with machine learning is that distinctions which do not seem important to us can have rather large effects on an algorithm's performance. 

\section{Benchmarking}

\subsection{Baseline tests and reproducibility}
We now present benchmarks of \textit{funsearch} that we performed using the three problems mentioned above. Specifically, we will evaluate the reproducibility of \textit{funsearch} for the cap-set, narrow-admissible-tuple, and no-isosceles problems, as shown in Tables \ref{tab:RepCapset}, \ref{tab:RepNAT}, and \ref{tab:RepIso}, respectively, across a range of different available models (LLMs). We also consider the effect of much longer \textit{funsearch} runs, using the gemini-2.0-flash model as an example, in Table.~\ref{tab:long}.

As a useful measure for these benchmarks, we define \textit{relative tokens} for a given model/provider via:
\begin{equation}
    n_R = \frac{p_O}{p_I} n_I + n_O,
\end{equation}
where $n_O$ and $n_I$ are the number of output and input tokens used, while $p_O$ and $p_I$ are the current prices of the tokens. All the benchmarks that we present here keep the number of relative tokens fixed, independent of the model, cost, or running time. The authors believe that this measure is less likely to be influenced by fluctuations in the cost per token or latency of available models.

Our first observation is that there is little correlation between the model's cost and its effectiveness for \textit{funsearch}. For instance, the most expensive model we tested, gpt4o, performed the best on average for the no-isosceles problem, was in the middle of the pack on the cap-set problem, and one of the worst performers on the narrow-admissible-tuple problem.

Additionally, as can be inferred from Table \ref{tab:long}, extended searches with \textit{funsearch} show little to no improvement.  Indeed, for all three of the problems we tested using the gemini-2.0-flash model, the best score achieved during 8 independent runs using $2\times 10^6$ relative tokens (total cost \$4) was substantially better than that achieved during a single run using $10^8$ relative tokens (total cost \$25).  One possible explanation for this is a path-dependence caused by LLMs being more influenced by their previous choices (which are included in the prompt) than is ideal; increasing the temperature might ameliorate this, but this might also decrease the performance at early stages of the run and reduce the proportion of viable responses (all the models will occasionally return responses that are not valid Python programs).

In Tables~\ref{tab:RepCapset}--\ref{tab:RepIso} below, each run used a $2\times10^6$ relative tokens, with 16 samplers, 16 evaluators, and an island reset time of 15 minutes. There is large variation in the cost of the open source models (indicated with a star) when accessed via open router, depending on the provider. We therefore state the average cost we paid for our benchmarks (run in February 2025).

\begin{table}[h]
\centering
\caption{Results for 8 runs of the \textbf{cap-set} problem with $n=8$ using $2\times10^6$ relative tokens in each run. The best known score is 512.}
\begin{tabular}{|l|c|c|c|c|c|}
\hline
\small{\textbf{model}}  & \small{\textbf{ave best}} & \small{\textbf{min best}} & \small{\textbf{max best}} & \small{\textbf{\#max}}& \small{\textbf{\$/run}} \\ \hline
mistral-tiny  & 387 & 366 & 448 & 1 & 0.8\phantom{$0^\ast$} \\ \hline
deepseek-chat & 368 & 354 & 394 & 1 & 3$^\ast$\phantom{$.00$} \\ \hline
gpt4o & 388 & 376 & 397 & 1 & 20\phantom{$.000^\ast$} \\ \hline
llama-3.3-70b-instruct & 375 & 354 & 392 & 1 & 0.88$^\ast$ \\ \hline
codestral-mamba & 390 & 364 & 448 & 2 & 0.5\phantom{$0^\ast$}\\ \hline
claude-3.5-haiku & 382 & 366 & 396 & 1 & 8\phantom{$.00^\ast$} \\ \hline
gemini-2.0-flash & 348 & 316 & 379 & 1 & 0.5\phantom{$0^\ast$} \\ \hline
\end{tabular}
\label{tab:RepCapset}
\end{table}

\medskip

\begin{table}[h]
\centering
\caption{Results for 8 runs of the \textbf{narrow-admissible-tuple} problem with $n=5000$ using $2\times10^6$ relative tokens in each run; best known score is 672.}
\begin{tabular}{|l|c|c|c|c|r|}
\hline
\small{\textbf{model}}  & \small{\textbf{ave best}} & \small{\textbf{min best}} & \small{\textbf{max best}} & \small{\textbf{\#max}}& \small{\textbf{\$/run}} \\ \hline
mistral-tiny & 558 & 532 & 575 & 1 & 0.8\phantom{$0^\ast$}\\ \hline
deepseek-chat & 558 & 421 & 561 & 4 & 3$^\ast$\phantom{$.00$} \\ \hline
gpt4o & 510 & 410 & 563 & 1 & 20\phantom{$.00^\ast$} \\ \hline
llama-3.3-70b-instruct & 541 & 499 & 561 & 4 & 0.88$^\ast$ \\ \hline
codestral-mamba & 510 & 469 & 560 & 1 & 0.5\phantom{$0^\ast$}\\ \hline
claude-3.5-haiku & 561 & 560 & 564 & 1 & 8\phantom{$.00^\ast$}\\ \hline
gemini-2.0-flash  & 503 & 434 & 582 & 1 & 0.5\phantom{$0^\ast$}\\ \hline
\end{tabular}
\label{tab:RepNAT}
\end{table}

\medskip

\begin{table}[h]
\centering
\caption{Results for 8 runs of the \textbf{no-isosceles} problem with $n=64$ using $2\times10^6$ relative tokens in each run; best known score is 110.}
\begin{tabular}{|l|c|c|c|c|r|}
\hline
\small{\textbf{model}}  & \small{\textbf{ave best}} & \small{\textbf{min best}} & \small{\textbf{max best}} & \small{\textbf{\#max}}& \small{\textbf{\$/run}} \\ \hline
mistral-tiny & 88.5 & 86 & 90 & 2 & 0.8\phantom{$0^\ast$}\\ \hline
deepseek-chat & 88.5 & 86 & 90 & 2 & 3$^\ast$\phantom{$.00$} \\ \hline
gpt4o & 89.3 & 86 & 96 & 1 & 20\phantom{$.00^\ast$} \\ \hline
llama-3.3-70b-instruct & 87.5 & 86 & 88 & 5 & 0.88$^\ast$ \\ \hline
codestral-mamba & 87.8 & 86 & 92 & 1 & 0.5\phantom{$0^\ast$}\\ \hline
claude-3.5-haiku & 88.8 & 88 & 90 & 3 & 8\phantom{$.00\ast$}\\ \hline
gemini-2.0-flash & 86.6 & 84 & 90 & 1 & 0.5\phantom{$0^\ast$}\\ \hline
\end{tabular}
\label{tab:RepIso}
\end{table}

\medskip

\begin{table}[h]
\centering
\caption{Results for single runs of gemini-2.0-flash on varying problems using $10^8$ relative tokens in each run. Each run cost \$25, and the islands reset every 15 minutes for the short reset time, and every hour for the long reset time. The change in best score as a function of relative tokens spent is shown in Figure~\ref{fig:LongBestChange} on a logarithmic scale.}
\begin{tabular}{| c | c | c | c |}
\hline
\small{\textbf{problem}}  & \small{\boldsymbol{$n$}} & \small{\textbf{short reset best}} & \small{\textbf{long reset best}} \\ \hline
cap-set & 8 & 370 & 349 \\ \hline
narrow-admissible-tuple & 5000 & 498 & 561 \\ \hline
no-isosceles & 64 & 90 & 86 \\ \hline
\end{tabular}
\label{tab:long}
\end{table}

\begin{figure} 
    \centering
    \begin{subfigure}{.5\textwidth} 
        \centering 
        \includegraphics[width=1\linewidth]{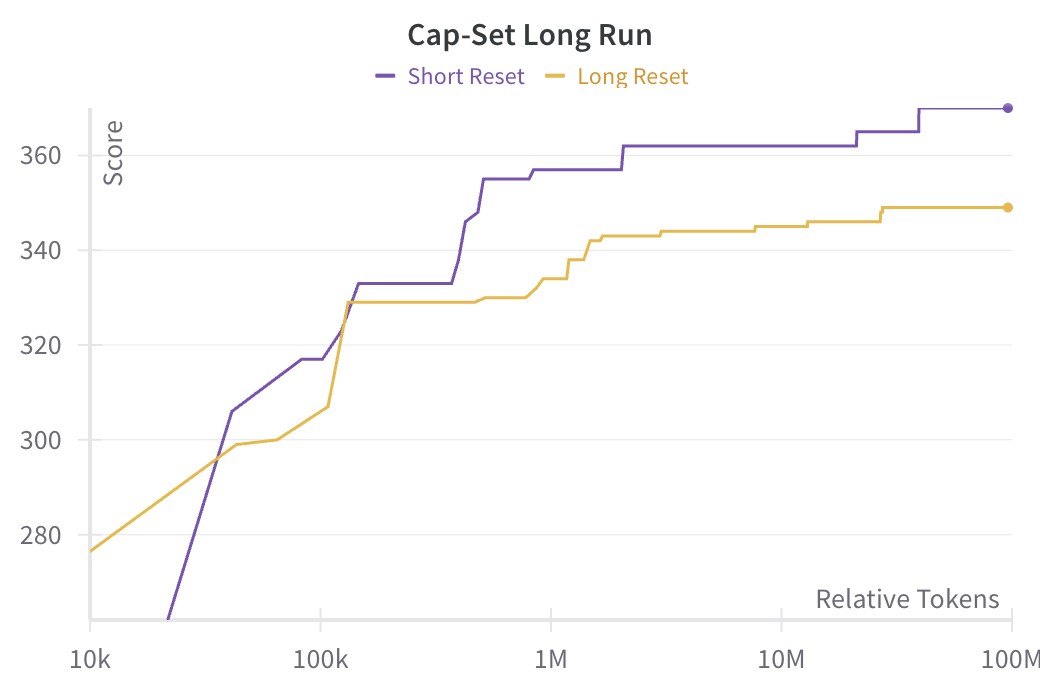} 
    \end{subfigure}%
    \begin{subfigure}{.5\textwidth} 
        \centering 
        \includegraphics[width=1\linewidth]{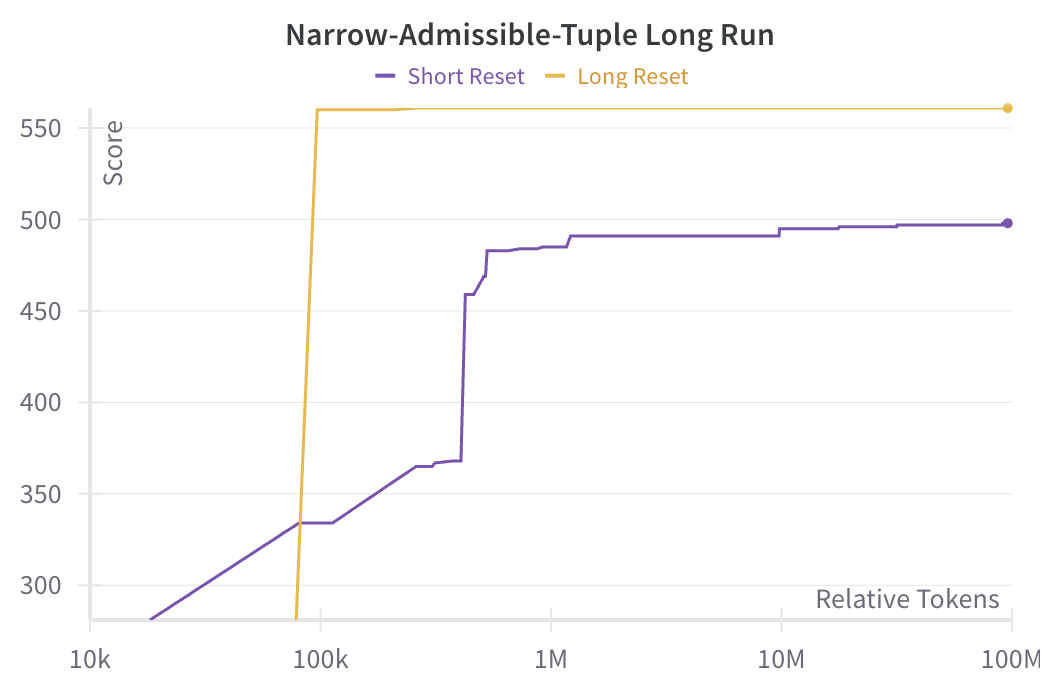} 
    \end{subfigure}%

    \begin{subfigure}{.5\textwidth} 
        \centering 
        \includegraphics[width=1\linewidth]{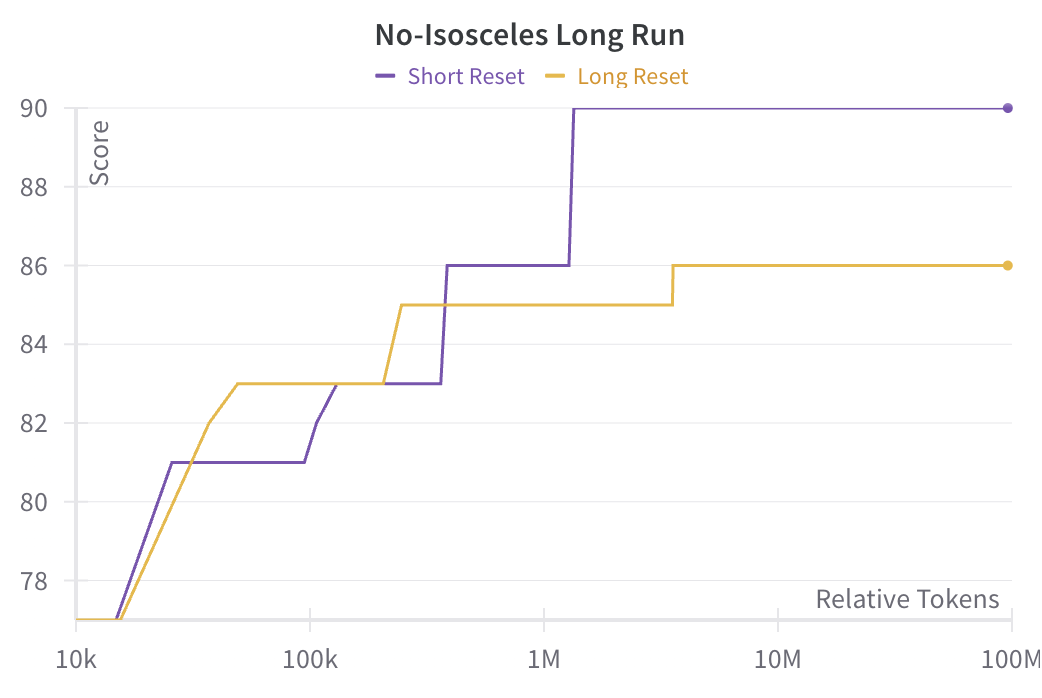} 
    \end{subfigure}
    
    \caption{The best score as a logarithmic function of relative token uses for the longer runs from Table~\ref{tab:long}.} 
    \label{fig:LongBestChange} 
\end{figure}


\subsection{Problem variants and generalization:\\experiments with no-isosceles}

The basic priority-solve setup we use for the no-isosceles problem is as follows:
\begin{itemize}
    \item {\it priority} takes as input an integer $n$ and a point on the $n \times n$ grid, and outputs a real number.
    \item {\it solve} takes $n$ as input, starts with an empty $n \times n$ grid, and adds points in order of {\it priority}, skipping any points that would create an isosceles triangle.  Terminate when no more points can be added without creating an isosceles triangle.  The {\it evaluate} function returns the size of the resulting set.
\end{itemize}

A fundamental difference between \textit{funsearch} and a more traditional machine learning protocol is that, {\it solve} being fixed and {\it priority} being learned, we are free to run the code on values of $n$ other than the value on which we trained, and see what kind of no-isosceles subsets of $n$ are produced.  In other words, we can ask to what extent training on a grid of one size {\em generalizes} to grids of a different size.

For that matter, we are also not restricted to training on grids of only one size. This is simply a matter of replacing {\it evaluate}; instead of returning the size of {\it solve}$(n)$ for a single $n$, we ask it to return the sum of the sizes of $({\textit{solve}}(n)/n)$ as $n$ ranges over some set of values or just $({\textit{solve}}(n)/n)$ when training on random values of $n$ in $[8,50]$. 

For the rest of this section, we will evaluate approaches based on training on individual values of $n$, a range of values of $n$, and on a random sampling of values of $n$. For models trained on individual $n$, we report only the models which performed the best in our experiments. All the models were run on mistral-tiny, with $10$ islands, $10$ samplers, and $8$ evaluators for $30$ minutes with the top priority functions were sampled at the end. 

\begin{figure}[t]
\captionsetup{font=small}
    \centering
    \begin{minipage}[t]{0.6\linewidth} 
        \centering        \includegraphics[width=\linewidth]{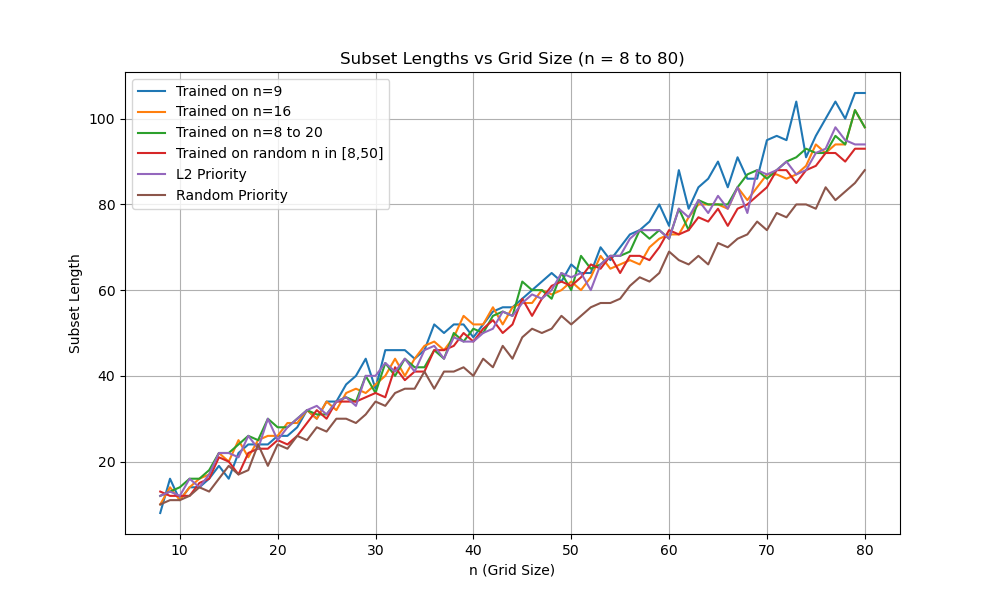}
        
        \caption{Graph of length of isosceles-free sets  generated by models trained on different values of n.}
        \label{fig:IsoGeneral}
    \end{minipage}
    \hfill
    \begin{minipage}[t]{0.35\linewidth} 
        \centering
        \includegraphics[width=\linewidth]{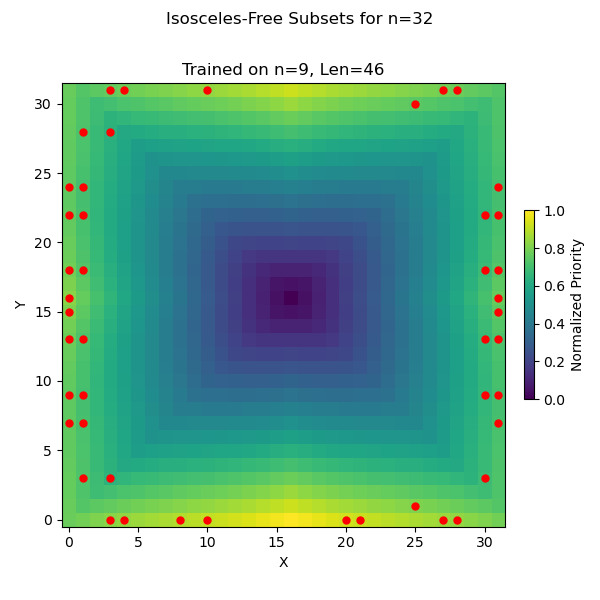}
        \caption{Isosceles-free subset of size $46$ for a $32 \times 32$ grid and heatmap representing the priorities assigned to each point by a model trained on $n=9$. The true largest subset has size $56$. }
        \label{fig:Iso9Basic32}
    \end{minipage}
\end{figure}

Whether we train on a single $n$ or on multiple values of $n$, we can run the resulting trained programs through {\it solve} for larger $n$ and record the size of the resulting no-isosceles sets. We record the results in Figure \ref{fig:IsoGeneral}

These priority functions were trained on the basic priority-solve setup described at the start of this section. Some remarks on these results:
\begin{itemize}
    \item The ``random priority'' line represents the result when {\em priority} assigns a random value to each point in the $n \times n$ grid.  This is a proxy for how well the greedy algorithm does without any training at all.
    \item All the trained priority functions outperform random priority, even for values of $n$ well outside the training range. This indicates that generalization between grid sizes is occurring.
    \item On the other hand, training on multiple values of $n$ does not seem to improve generalization.  One might have thought multiple-$n$ training would differentially favor attributes of priority functions that enhance performance in ways that don't depend on $n$; but we did not find any such effect in this experiment.
    \item None of our basic priority functions are generating optimal no-isosceles sets, even on the $n$ for which they are trained, for $n \geq 12$.  For $n$ up to around $32$, the size of the largest no-isosceles set can be computed exactly by SAT solvers.  The largest no-isosceles set in the $16 \times 16$ grid has size $28$, while the largest set found by the basic model trained on $n=16$ had size $26$. (But see Table \ref{tab:SymVsBasic} below: modifying {\em it solve} can extent the range in which funsearch achieves best possible results.) For $n$ outside this range, our \textit{funsearch} experiments do not achieve results as strong as those of PatternBoost~\cite{patternboost}.  Note, though, that we are devoting substantially less computational resources to each experiment than were used in that paper.
\end{itemize}

Direct examination of the priority function helps illuminate the results above.

Figure~\ref{fig:Iso9Basic32} shows the priority function on the $32\times 32$ grid learned by training on $n=9$, together with a no-isosceles set obtained by running {\it solve} with that priority.  The priority function visibly favors points near the edge of the grid; indeed, trained priority functions almost invariably have this feature, no matter what size or sizes of grids they train on, and produce no-isosceles sets in which most of the points are near the boundary.  In fact, every machine learning approach we have tried on the no-isosceles problem yields point sets which favor the boundary.  With a little thought, one can make sense of this.  The condition that a triangle $P,Q,R$ is isosceles is given by the vanishing of a quadratic function in the coordinates of the integer vectors $Q-P$ and $Q-R$.  The larger these coordinates, the smaller the heuristic probability that this quadratic function will evaluate to $0$.  And so it's reasonable that placing the points far away from each other is helpful in reducing the number of triangles, and arranging the points near the boundary of the grid tends to increase those pairwise distances.  This is obvious once observed, but we will admit it did not occur to us until we saw what outputs of machine approaches to the problem looked like.  Figure~\ref{fig:Iso9Basic32} also includes scores for the hand-built priority function that assigns points their distance from the center to the center under an $l_2$ norm. We can see in the plot that while the $l_2$ priority function does also perform better than the random model, the model trained on $n=9$ seems to generalize slightly better still. 

 It seems possible that the no-isosceles problem splits into a generalizable part (tactics that work well for all $n$, or at least an infinite class of $n$) and a non-generalizable part (tactics that are specific to an individual $n$ or finite range of $n$.) 
 In the no-isosceles, training on even one $n$ is sufficient to capture a tactic in the generalizable part.  One then might ask: is anything about the no-isosceles problem generalizable besides the importance of staying near the boundary?  The apparent slight superiority of the $n=9$ trained priority to $l_2$ priority suggests a positive answer.

 \begin{figure}[t]
 \captionsetup{font=small}
    \centering
    \begin{minipage}[t]{0.6\linewidth} 
        \centering        \includegraphics[width=\linewidth]{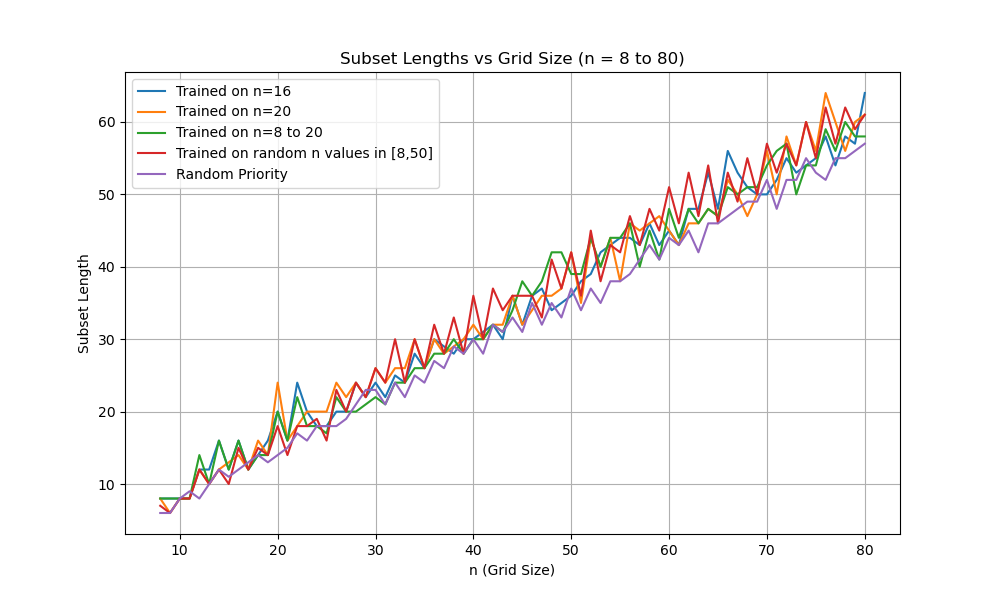}
        \caption{Graph of length of isosceles-free sets on a lattice embedded on a torus generated by models trained on different values of n. }
        \label{fig:TorGeneralization}
    \end{minipage}
    \hfill
    \begin{minipage}[t]{0.35\linewidth} 
        \centering
        \includegraphics[width=\linewidth]{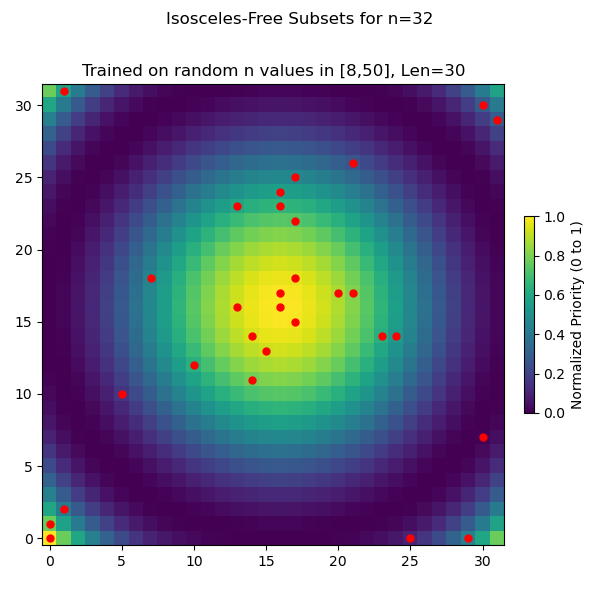}
        \caption{Isosceles-free subset of size $30$ for a $32 \times 32$ grid embedded on a torus and heatmap representing the priorities assigned to each point.}
        \label{fig:TorRandN32}
    \end{minipage}
\end{figure}

One way to approach this question is to consider a variant of the problem, in which we search for subsets of the {\em torus} $(\mathbf{Z}/N\mathbf{Z}) \times (\mathbf{Z}/N\mathbf{Z})$ with no isosceles triangles.  In other words, we connect the opposite edges of the grid to form a surface with no boundary; in so doing, we eliminate the one generalizable tactic we know about.  The results of training \textit{priority} on small $n$ and testing on larger $n$ are shown in Figure~\ref{fig:TorGeneralization}. We see that the advantage of trained models over random priority is much reduced; but it is still visible. \textit{Something} is being learned by the model trained on $n=9$ that generalizes to much larger grids.  But we are not sure what it is.  The priority function (Figure~\ref{fig:TorRandN32}) seems to be clustered around the center. The best generalizing model here was also the trained on random values of $n$ in $[0,50] $ and unlike the previous case, training on more values of $n$ does improve the generalizability of the model outside its training range. 

The separation between {\it evaluate}, {\it solve}, and {\it priority} gives the researcher a great deal of flexibility in modifying the problem, or \textit{funsearch}'s approach to the problem; this is very useful in a context like this one where we have very little intuition about what approach is likely to be effective.

 \begin{figure}[t]
 \captionsetup{font=small}
    \centering
    \begin{minipage}[t]{0.6\linewidth} 
        \centering        \includegraphics[width=\linewidth]{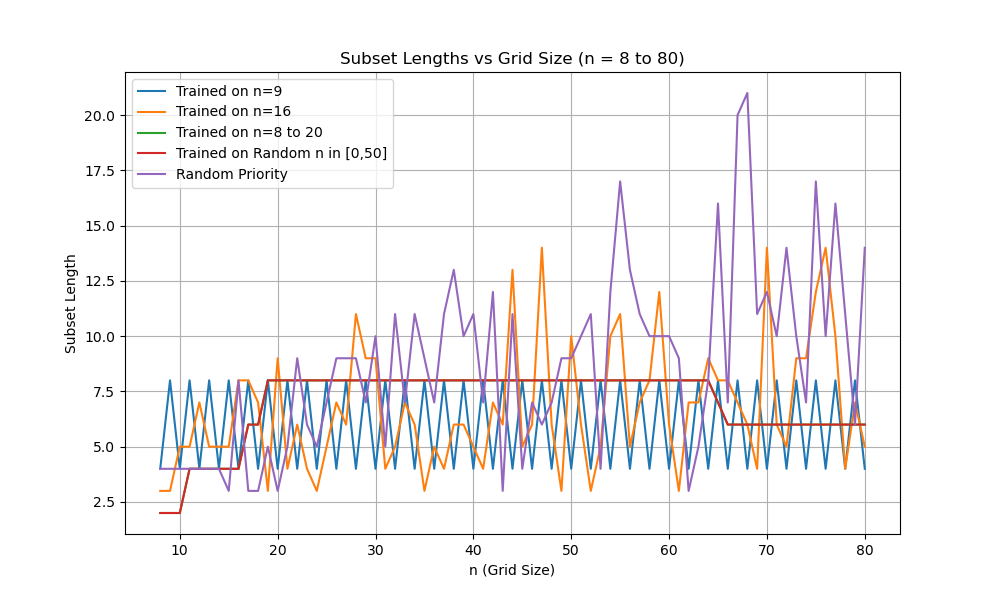}
        \caption{Size of isosceles-free subsets when having priority decide which points to remove rather than add. We see that on average, randomly choosing points to remove is a better protocol than the learned ones.}
        \label{fig:RevGeneralization}
    \end{minipage}
    \hfill
    \begin{minipage}[t]{0.35\linewidth} 
        \centering
        \includegraphics[width=\linewidth]{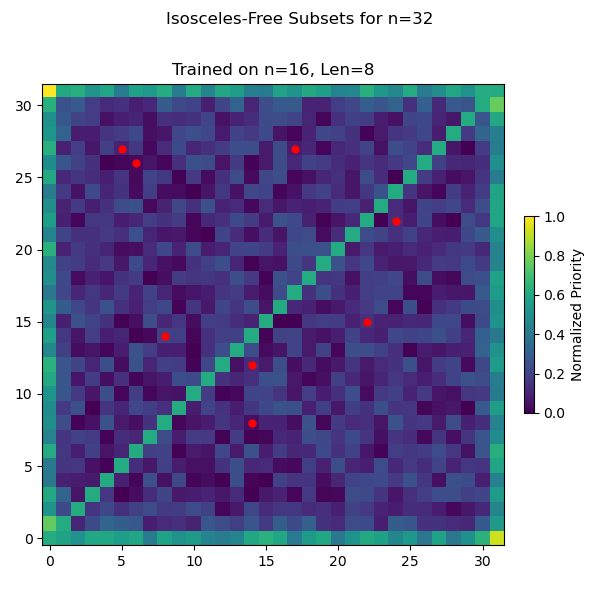}
        \caption{Isosceles-free subset of size $8$ for a $32 \times 32$ grid generated by removing points chosen by a model trained on $n=16$. High priority indicates points to be removed here.}
        \label{fig:Rev32}
    \end{minipage}
\end{figure}

For instance, instead of having solve start with the empty grid and add points in priority order, skipping any that would create an isosceles triangle, we could start with every point in the grid included, and {\em remove} points in order of priority, stopping as soon as we have a subset that is isosceles free. The results of evolving priority with this solve function are shown in Figure~\ref{fig:RevGeneralization}, with a specific example of a model trained on $n=16$ shown in Figure \ref{fig:Rev32} for a $32 \times 32$ lattice. Notice that this set is not maximal! There are many points that can be added to this set while remaining isosceles free. These points were prioritized to be removed earlier and therefore do not survive. We see that although the problem of prioritizing which points to add as opposed to which to remove is mathematically equivalent in formulation, we get significantly different results. 

\begin{figure}[t]
 \captionsetup{font=small}
    \centering
    \begin{minipage}[t]{0.6\linewidth} 
        \centering        \includegraphics[width=\linewidth]{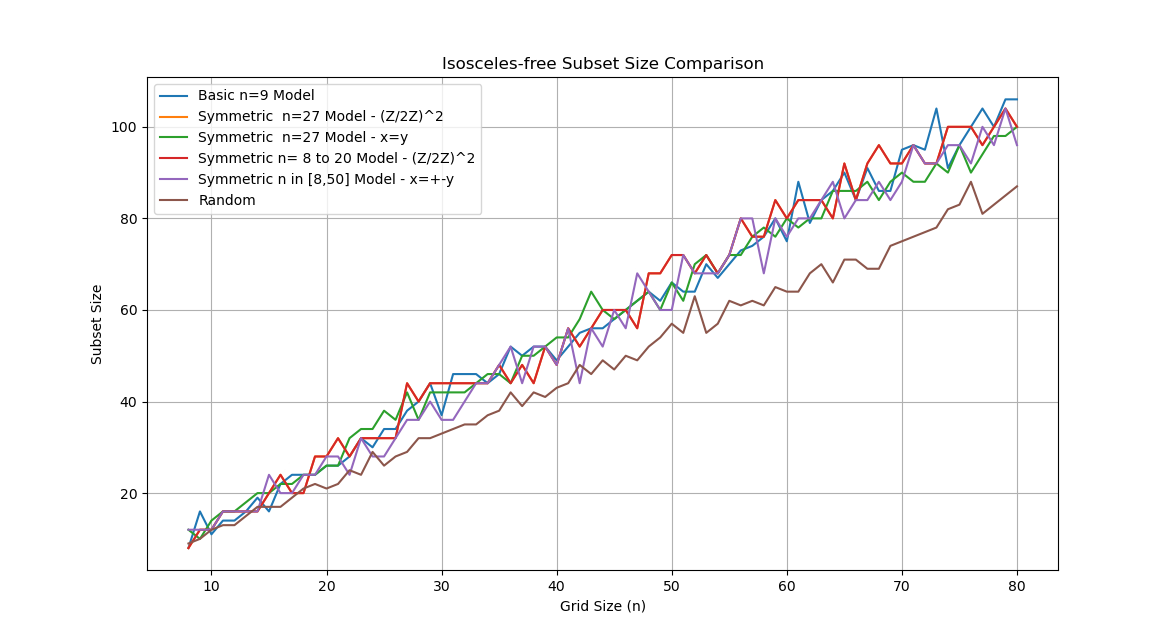}
        \caption{Generalization graph for various models trained on both different values of n and different symmetries. }
        \label{fig:SymGen}
    \end{minipage}
    \hfill
    \begin{minipage}[t]{0.35\linewidth} 
        \centering
        \includegraphics[width=\linewidth]{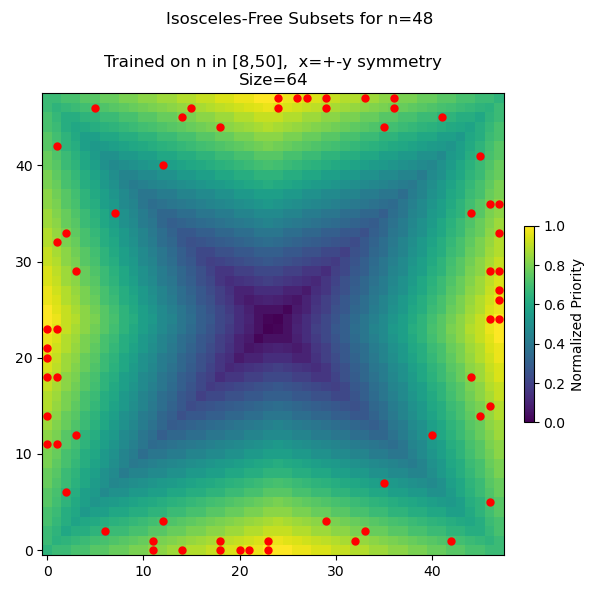}
        \caption{Isosceles-free subset of size $64$ for a $48 \times 48$ grid with symmetry over the lines $x=\pm y$ }
        \label{fig:Sym48}
    \end{minipage}
\end{figure}

In other experiments with no-isosceles, it has sometimes been found to be useful to impose on the learned solution some group of symmetries (in particular some subgroup of the group of symmetries of the problem, which is dihedral of order $8$.) Again, this is easy to implement in \textit{funsearch}.  As an example, we impose a two different symmetry groups:  $(\mathbb{Z}/2\mathbb{Z})^2$ and $\mathbb{Z}/2\mathbb{Z}$. We then assign priorities like in the basic model to only a part of the grid and solve adds one {\em orbit} of points at a time in order of priority, skipping any that would create isosceles triangles. We tested on grids that are symmetric under the following transformations: 
\begin{itemize}
    \item flips over the central $x$ and $y$ axes - $(\mathbb{Z}/2\mathbb{Z})^2$ symmetric symmetry.
    \item flips over the line $x=y$ -  $\mathbb{Z}/2\mathbb{Z}$ symmetric.
    \item  flips across both $x=y$ and $x=-y$.
\end{itemize}
The generalization results for all of these models are shown in Figure~\ref{fig:SymGen} and a specific example shown in Figure~\ref{fig:Sym48}. We compare this to the best generalizing model from the basic functions as well as a priority function that assigns random values. As we see, adding symmetry to the mix improved generalization across all models and the generalization is comparable to the best basic model trained on $n=9$. 

While the generalization graphs show that the best models we have found so far have similar generalizability, it is important to remember that this represents the performance of each priority function outside its training data. If we train on different values of $n$ and look at the performance of priority on the trained values, we see a difference in performance between the basic and symmetric models, as seen in Table \ref{tab:SymVsBasic}.  
\begin{table}[h]
\centering
\caption{Comparison of Basic Models and Symmetric Model}
\begin{tabular}{| l | r | r | r | r | r | r | r | r | r |}
\hline
\multirow{2}{*}{\textbf{problem setup}} & \multicolumn{9}{c|}{{$\boldsymbol{n}$}} \\ \cline{2-10}
 & \textbf{12} & \textbf{13} & \textbf{16} & \textbf{21} & \textbf{23} & \textbf{25} & \textbf{27} & \textbf{32} & \textbf{64}\phantom{$^*$} \\ \hline
maximum known & 20 & 22 & 28 & 36 & 40 & 44 & 48 & 56 & 110\phantom{$^*$}  \\ \hline
basic models & 20 & 20 & 26 & 34 & 36 & 40 & 40 & 46 & 86$^*$  \\ \hline
symmetric models & 20 & 22 & 28 & 36 & 40 & 40 & 44 & 52 & 96\phantom{$^*$} \\ \hline
\end{tabular}
\label{tab:SymVsBasic}
\end{table}

We see the symmetric models perform slightly better on the values they have been trained on. \textsuperscript{*}We also note that the $86$ achieved by the basic model can be improved to $96$ with more compute as seen in Table~\ref{tab:RepIso}, but setting up the problem symmetrically saves on computational cost.

 \begin{figure}[t]
 \captionsetup{font=small}
    \centering
    \begin{minipage}[t]{0.6\linewidth} 
        \centering        \includegraphics[width=\linewidth]{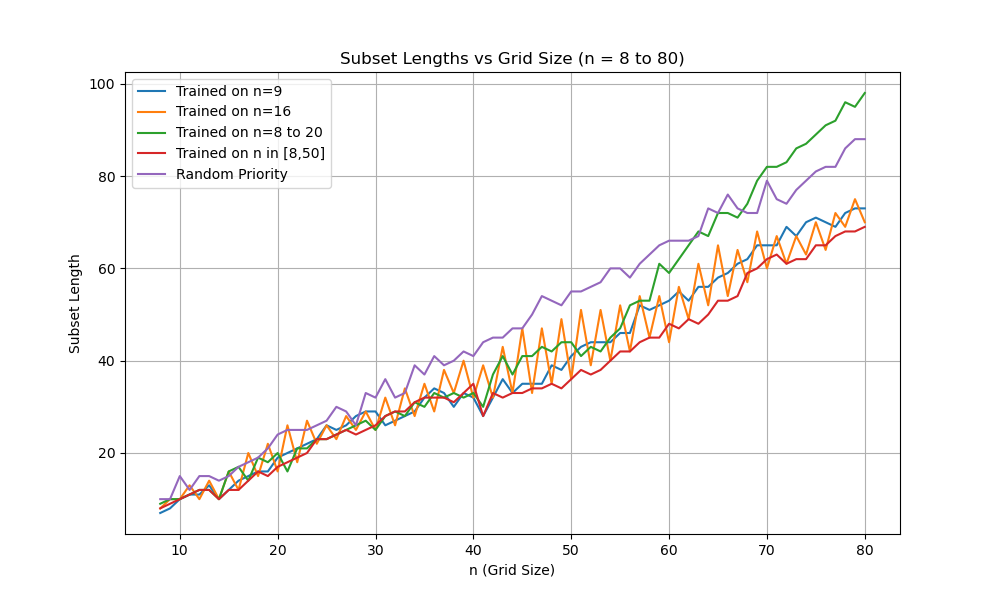}
        \caption{Size of small maximal isosceles-free subsets of an $n \times n$ lattice vs $n$. Points added in order of priority but preference given to smaller sets - being further below random is better here.}
        \label{fig:SmallMaxGen}
    \end{minipage}
    \hfill
    \begin{minipage}[t]{0.35\linewidth} 
        \centering
        \includegraphics[width=\linewidth]{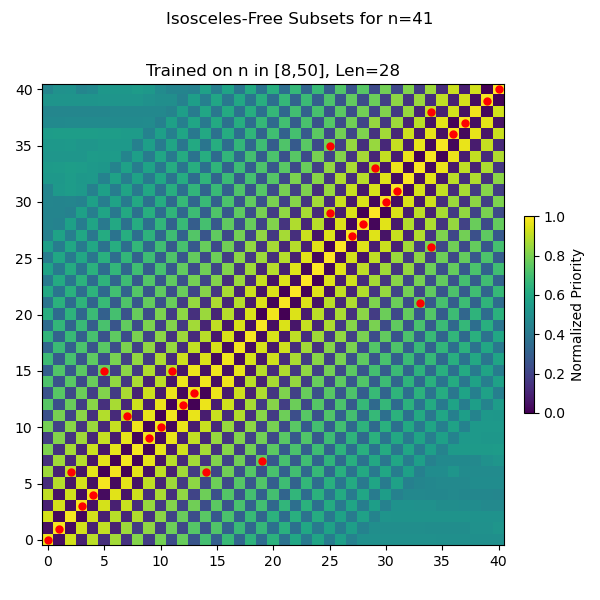}
        \caption{Small maximal isosceles-free subset of size $28$ for a $41 \times 41$ grid.}
        \label{fig:SmallMax41}
    \end{minipage}
\end{figure}

A very small change can create a whole new experiment of interest.  For instance, suppose we evaluate by the negative of the size of the set returned by {\it solve}, instead of its size.  That is, we reward priority functions which result in an output of {\it solve} (necessarily a maximal no-isosceles set) which are as {\em small} as possible.  The smallest such sets found by \textit{funsearch} are shown in Figure~\ref{fig:SmallMaxGen} with a particular example of a good generation shown in Figure~\ref{fig:SmallMax41}. Note that the model trained on $n=16$ is sensitive to parity, with substantially better performance for even $n$; when studying generalization it is probably best to make sure the test parameters don't all lie in an arithmetic progression.
  The problem of small maximal no-isosceles subsets has not been considered before, as far as we know, and it may be that human effort could improve these bounds without much difficulty.  The analogous problem for no-three-in-line (finding small subsets of the grid with no three in a line, and which cannot be enlarged without violating this condition) is, on the other hand, an old one~\cite{geometricdominating}.

\begin{figure}[t]
 \captionsetup{font=small}
    \centering
    \begin{minipage}[t]{0.6\linewidth} 
        \centering        \includegraphics[width=\linewidth]{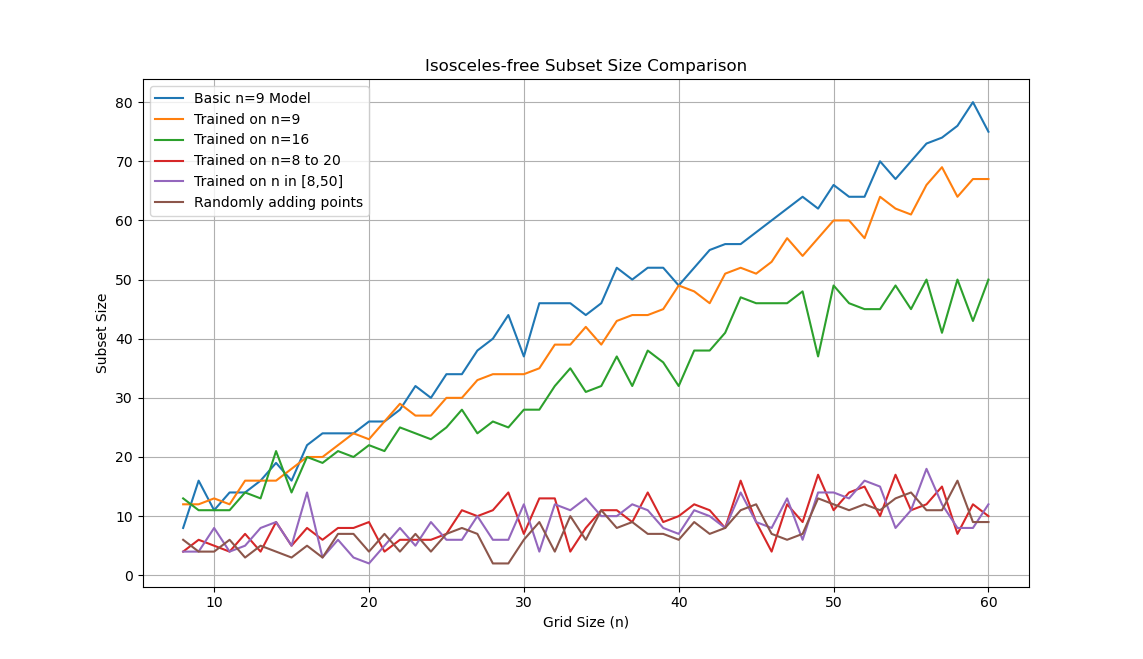}
        \caption{Size of isosceles-free subsets of an $n \times n$ lattice when predicting the next point instead of assigning priorities to each point.}
        \label{fig:NextPointGenGen}
    \end{minipage}
    \hfill
    \begin{minipage}[t]{0.35\linewidth} 
        \centering
        \includegraphics[width=\linewidth]{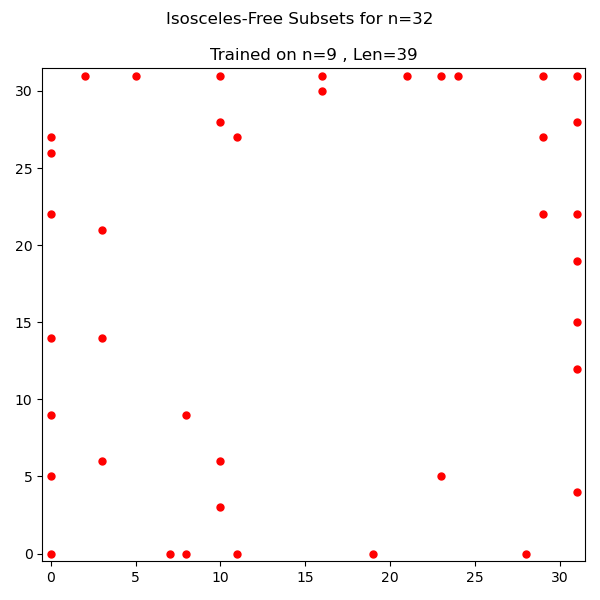}
        \caption{Large isosceles-free subset of size $39$ for a $32 \times 32$ grid. No heatmap since the priority function is directly predicting the points!}
        \label{fig:NextPoint32}
    \end{minipage}
\end{figure}

Finally:  one surprising feature of the baseline \textit{funsearch} setup is that the priority function is determined in advance.  That is, when we choose which grid point to add, we are not taking into account which points we've already added, except insofar as we forbid isosceles creation.  This is very strange!  A chess engine that ranked all possible moves in advance, and then from each board state played the highest-priority move legal from that position, would play very bad chess.  Changing this feature is, again, easy.  We recast {\it priority} so that, instead of taking a grid point as input and returning a number, it takes a subset of the grid as input and returns a point.  Then {\it solve}, at each step, adds the point chosen by priority, skipping over suggested points that would result in isosceles triangles. Given that, in theory, priority can repeat suggestions or suggest points outside the lattice, we have a budget for the number of suggestions priority is allowed to make. One could think that given the trends observed with the best subsets generated so far, the largest subset grows on the order of $O(n)$ and specifically around $\sim 1.7n$. So we first keep a budget of $3n$ as an upper bound but find that we get a better result with a budget of on the order $O(n^2)$. The generalization of these models is shown in Figure~\ref{fig:NextPointGenGen} with a particular example in Figure~\ref{fig:NextPoint32}. These models have a wider range of performance, but perhaps counter-intuitively still do worse than the basic models!

Overall, we tested four methods of generating isosceles free sets: with the basic \textit{solve}-\textit{priority} setup, symmetric generation, removing points from a full grid, and suggesting points directly instead of assigning priorities. We saw that we get different results from mathematically equivalent formulations of the same problem. And we saw that with small changes to the code, we can test different problems entirely --- embedding the isosceles-free subset problem onto a torus as well as looking for small, maximal isosceles-free sets.

\section{Conclusion}
In this article, we benchmarked \textit{funsearch} against a variety of mathematical problems, specifically the cap-set, narrow-admissible-tuple, and no-isosceles problems. One key insight that emerged is that models with different costs sometimes yielded similar results, underscoring the importance of using multiple models for comprehensive searches. It can also be observed that little is gained from extended searches with \textit{funsearch}. It appears far more beneficial to run a large number of shorter searches.  In the context of the no-isosceles problem, we demonstrated that \textit{funsearch} can be easily deployed on many variants of the same problem, and in fact can be deployed on different presentations of the same problem, often with very different performance.

We also provide an implementation of \textit{funsearch} in the GitHub repository~\cite{funsearch}. Building on the original design, we add support for searches beyond the traditional priority-function framework, enabling searches for programs with arbitrary type signature. We also add the ability to use multiple models simultaneously, or easily switch between models using Openrouter. Finally, the repository also contains many quality-of-life improvements, such as Weights-and-Biases integration, which makes monitoring progress much more straightforward.

The authors believe that the work, and implementation of \textit{funsearch}, presented here, demonstrate that LLM-driven genetic methods like \textit{funsearch} can be brought to bear on a wide range of mathematical problems of interest at only modest cost and without any requirement that mathematicians retrain in implementation of machine learning methods. We hope this will be beneficial to mathematicians seeking innovative computational solutions for mathematical discovery.

\section*{Acknowledgments}
This project was started at the Harvard CMSA Mathematics and Machine Learning Program held in fall 2024.
The authors would like to thank Mistral AI for generously providing credits for their large language models. CSFT would like to acknowledge the hospitality of IAIFI at the Massachusetts Institute of Technology, where a portion of this research took place. 

CSFT is supported by the Gould-Watson Scholarship. TRH is supported by the National Science Foundation under Cooperative Agreement PHY-2019786 (The NSF AI Institute for Artificial Intelligence and Fundamental Interactions, http://iaifi.org/). AVS is supported by Simons Foundation grant 550033.  JSE is supported by NSF grant DMS-2301386, and by the Office of the Vice Chancellor for Research and Graduate Education at the University of Wisconsin-Madison with funding from the Wisconsin Alumni Research Foundation.  Some of the computations in this paper were run on the FASRC Cannon cluster supported by the FAS Division of Science Research Computing Group at Harvard University.

\appendix

\section{Example Specification}\label{app:ExampleSpec}
\begin{lstlisting}[language=Python]
### SYSTEM PROMPT
"""<<<system prompt can be specified here>>>"""
### END SYSTEM PROMPT
"""Finds sets.
On every iteration, improve the priority_v# function over
the priority_v# methods from previous iterations.
Make only small changes. Try to make the code short.
"""

import itertools
import numpy as np
import funsearch
import math

@funsearch.run
def evaluate(n: int) -> int:
  return solve(n)

def solve(n: int) -> int:
  """Returns the count of numbers between 5 
  and n-1 where is_prime(m) matches (m)."""
  final_count = 0
  for m in range(5, n):
      is_prime = True
      if m <= 1:
          is_prime = False
      else:
          for i in range(2, int(math.sqrt(m)) + 1):
              if m % i == 0:
                  is_prime = False
                  break
      pr = priority(m)>0.5
      if is_prime == pr:
          final_count = final_count + 1
  return final_count
  

@funsearch.evolve
def priority(n: int) -> bool:
  """Returns 1 if add to set, 0 otherwise.
  n is an int.
  """
  return True
\end{lstlisting}

The system prompt can optionally be specified at the start of the specification file. If it is not specified, the default in \texttt{config.py} is be used. If the default is used, the system prompt will be adjusted to describe whichever function \texttt{@funsearch.evolve} decorates. The API prompt constructed from this specification file then looks like:

\begin{lstlisting}[language=Python]
=== SYSTEM PROMPT ===
<<<specified system prompt>>>
=== END SYSTEM PROMPT ===

=== PROMPT ===
"""Finds sets.

On every iteration, improve the priority_v# function over 
the priority_v# methods from previous iterations.
Make only small changes.
Try to make the code short.
"""

import itertools
import numpy as np
import funsearch
import math

@funsearch.run
def priority_v0(n: int) -> bool:
  """Returns 1 if add to set, 0 otherwise.
      n is an int.
  """
  return True


def priority_v1(n: int) -> bool:
  """Improved version of `priority_v0`.
  """
=== END PROMPT ===
\end{lstlisting}
\bibliography{bibliography.bib}
\bibliographystyle{inspire}
\end{document}